\documentclass{article}
\usepackage{spconf,amsmath,graphicx}
\usepackage{amsthm,amsfonts,amssymb}
\usepackage[export]{adjustbox}
\usepackage{xcolor}
\usepackage[linesnumbered,ruled,vlined]{algorithm2e}

\graphicspath{{pics/}}

\SetCommentSty{mycommfont}

\SetKwInput{KwInput}{Input}                
\SetKwInput{KwOutput}{Output}              
\SetKwInput{KwModel}{Model Used}

\newcommand{\argmax}{\operatornamewithlimits{argmax}}

\def\RR{\mathbb R}

\newcommand{\bp}{\mathbf{p}}
\newcommand{\bq}{\mathbf{q}}

\newcommand{\bx}{\mathbf{x}}

\title{Latent Image and Video Resolution Prediction using Convolutional Neural Networks}
\name{Rittwika Kansabanik, Adrian Barbu 
}
\address{Statistics Department, Florida State University}
\begin{document}
%
\maketitle
\begin{abstract}
This paper introduces a Video Quality Assessment (VQA) problem that has received little attention in the literature, called the latent resolution prediction problem. 
The problem arises when images or videos are upscaled from their native resolution and are reported as having a higher resolution than their native resolution. 
This paper formulates the problem, constructs a dataset for training and evaluation, and introduces several machine learning algorithms, including two Convolutional Neural Networks (CNNs), to address this problem.
Experiments indicate that some proposed methods can predict the latent video resolution with about $95\%$ accuracy. 
\end{abstract}
\vspace{-3mm}
\section{Introduction}
\vspace{-1mm}

\label{sec:intro}
In recent years, multimedia technologies have advanced massively, causing an explosion of digital visual content. 
According to the Cisco® Visual Networking Index (VNI), nearly $650 \times 10^6$ mobile devices and connections were newly added. 
The forecast says that mobile data traffic will grow by $46\%$ in the next year. 
This massive growth in the use of smart devices has caused tremendous exposure to images and videos to the human eye. Therefore, ensuring the quality of the end-user experience is very important. Many factors, including transmission rate and compression factors, can affect the perceived quality of an image or video. 
As the bandwidth of internet connections increases, this paper assumes a sufficiently high transmission rate and considers the compression artifacts minimal. 

\begin{figure}[t]
\centering
\fbox{\includegraphics[width = 8cm]{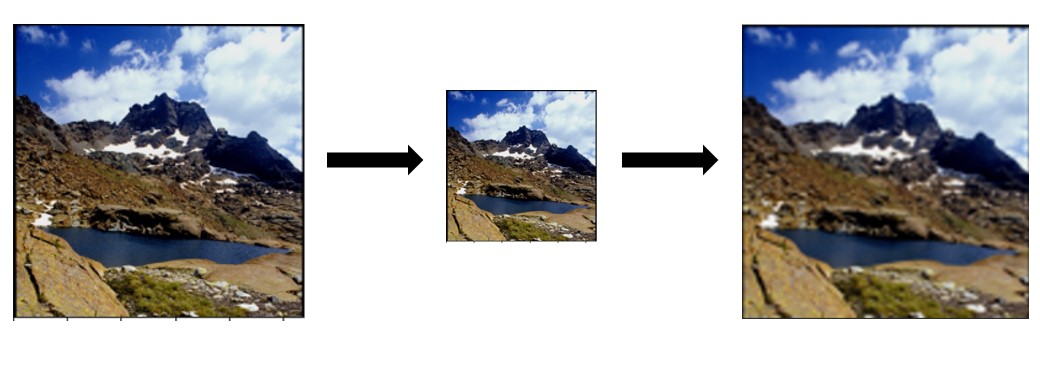}}
\vskip -3mm
\caption{
(Illustration) An image or video is claimed to be of a specific resolution (right), but it is an up-scaled version of a lower-resolution image (middle). To simplify the problem, we start with a high-resolution image (left), which is down-scaled and then up-scaled back to the original resolution. 
The problem is to predict the downscale/upscale factor used.}
\label{fig:latentres}
\vspace{-5mm}
\end{figure}

Nowadays, people upload images and videos on social media platforms. These contents are often claimed to be of high resolution (e.g., 1080p), which requires more disk space to store. 
In reality, many are of lower resolution contents (e.g., 480p), upscaled to the claimed resolution, as illustrated in Figure \ref{fig:latentres}. 
This work aims to build algorithms capable of predicting the latent resolution of images or videos and verifying that they are of the claimed resolution, improving user experience. 

This paper brings the following contributions:\vspace{-2mm}
\begin{itemize}
\item It introduces the problem of latent resolution prediction, which has not received any attention in the Video Quality Assessment literature.\vspace{-2mm}
\item It introduces two Deep Learning (DL) approaches based on Convolutional Neural Networks (CNN) in regression and classification settings to predict the latent resolution in images and videos. The Mask-CNN propagates a mask to keep track of the corner locations and obtain predictions from informative locations in the image. 
The SoftMax CNN makes predictions from multiple patches and uses percentiles to obtain a unified prediction for the image. \vspace{-2mm}
\item It conducts experiments on a dataset created for this task, containing images and videos with different latent resolutions ranging from 144 to 1080, all up-scaled to 1080. 
The experiments indicate that the proposed methods can estimate this scenario's latent image/video resolution with high accuracy.
\end{itemize}

\vspace{-5mm}
\subsection{Related Work}

This work is part of the more extensive Video Quality Assessment (VQA) area. 
The human visual system (HVS) is the most reliable way to assess perceived video quality. Several researches have been conducted to replicate the HVS \cite{lee2006objective}, but they could not develop any valuable measures. 
The most reliable source of quality assessment is based on human opinions \cite{aflaki2015subjective}, which subjective experiments can obtain. 
These experiments are time-consuming, expensive, and highly depend upon the individual’s mental and emotional state and physical conditions \cite{scott2016personality}.
Therefore, objective VQA measures became essential.

Based on the availability of reference videos, VQA methods can be divided into three groups: full-reference (FR-VQA), reduced-reference (RR-VQA), and no-reference (NR-VQA).

FR-VQA assumes that the original video is available as a reference.
The Peak Signal-to-Noise Ratio (PSNR) has been mainly used to evaluate the quality of video signals. 
It uses the Mean Square Error (MSE) between each frame of the reference and the processed video signal \cite{winkler2008evolution}.
Other well-known approaches include the Structural Similarity index (SSIM) \cite{wang2004video}, which can be extended to Multi-Scale (MS-SSIM)\cite{wang2003multiscale}, and the Motion-based Video Integrity Evaluation (MOVIE)\cite{seshadrinathan2009motion}. 

NR-VQA does not require access to the original video at all. Some early work in this field includes video BLINDS, proposed by the authors in \cite{saad2014blind}, which combines Discreet Cosine models with a motion-based model.
ML-based methods have been recently proposed to predict the perceived quality\cite{narwaria2011svd}. 
ML methods typically rely on two steps: feature extraction, in which representative features of the video content are computed, and classification/regression, where the extracted features are mapped into class scores or the training algorithm directly predicts the quality value.
Feature extraction is the most crucial part of a typical ML system. 
If the features extracted from the data are poor, these models fail to produce good results. 
After the massive success of CORNIA \cite{ye2012unsupervised}, in \cite{xu2014no} were extracted frame-level features via unsupervised feature learning, and a support vector regressor (SVR) was applied to map these onto subjective quality scores.  
Similarly, Video Intrinsic Integrity and Distortion Evaluation Oracle (VIIDEO) \cite{mittal2015completely} does not require human ratings
on video quality. It is developed based on the assumption that the normalized version of frame level differences will follow Gaussian distribution for good-quality videos.
Another method has been proposed in\cite{yan2018no}. It involves extracting perceptual features containing a more comprehensive range of spatiotemporal information from multidirectional video spatiotemporal slice (STS) images and uses a support vector machine (SVM) to evaluate video quality. 

Due to the subjective nature of this problem, it is not easy to define a set of features that appropriately quantify the actual mechanism of VQA. 
Deep learning (DL) models can acquire remarkable generalization capabilities when sufficient data is used for training. Unlike traditional machine learning techniques, they do not depend on sophisticated feature extraction and selection techniques. 
A deep unsupervised learning scheme was proposed in \cite{vega2017deep} based on noise ratio and motion intensity. 
DL models based on Convolutional Neural Networks (CNNs) have recently been used for picture-quality prediction \cite{kim2017deep}. 
In \cite{giannopoulos2018convolutional}, the authors proposed a Deep Learning (DL) framework that deploys CNNs to deal with compression distortion and transmission delays. 
In recent development, LSTM has also been applied along with CNN for VQA
In \cite{varga2019no}, a method based on transfer learning was introduced, where pre-trained CNNs like Inception-V3\cite{szegedy2016rethinking} and AlexNet\cite{krizhevsky2012imagenet} were used for feature extraction and an LSTM (Long Short-Term Memory) model was used for quality prediction.

Unlike the above works, this paper deals with the actual resolutions of the images. 
Images and videos can contain several kinds of distortions, and it is difficult to develop universal models that predict image quality scores in the presence of multiple distortions of the visual content. 
Therefore, this paper focuses on a specific kind of distortion that has received little attention in the past: the loss in quality due to the upscaling of images and videos to a higher resolution than their native resolution. 

\vspace{-3mm}
\section{Method Description}
\label{sec:method}
\vspace{-1mm}

This paper introduces two CNN-based methods for latent resolution prediction. 
Both methods focus on the textured parts of the image but differ in how they handle the image.
The first one uses image patches, extracted from the locations of interest in the image.
The second is a fully convolutional network that takes the whole image as input and produces an output map. 
At the same time, it propagates a binary mask containing the interest point locations to keep track of their location in the output map so that reliable quality predictions are extracted only from the corresponding locations. 

\vspace{-2mm}
\subsection{The Latent Resolution Problem} \label{sec:latent}
\vspace{-1mm}

As discussed in the introduction, the latent resolution is the factor used to upscale a lower-resolution image to obtain a given image $I$.
For better tractability, as shown in Figure \ref{fig:latentres}, we start with a high-resolution image that is down-scaled by a specific factor $k\in (0,1]$ and then is up-scaled by $1/k$ to obtain an image of the exact resolution as the original image. 
This way, the downscale/upscale factor is known, and machine learning models can be trained to predict it.

The problem is to predict $k$. 
Two versions of the problem will be considered: a regression version where the latent resolution $k =a/100$ with $a \in \{1,2,...,100\}$, and a classification version where $k = a/1080$, with $a \in \{144,240,360,420,720,1080\}$.

\vspace{-3mm}
\subsection{Interest Points}
\vspace{-1mm}

Not all parts of the image are equally crucial for predicting the latent resolution. 
Flat areas are almost identical in high- and low-resolution images, and the differences between different resolutions are noticeable only in textured regions. 
Textured areas can be identified using corner detection. 
In this paper, the corners are detected using the standard Harris corner detector \cite{harris1988combined}. 
Non-maximal suppression with a radius of ten is used to obtain a few non-overlapping corners across the image. 

\vspace{-3mm}
\subsection{Patch-Based CNN}
\vspace{-1mm}
\noindent{\bf Patch Extraction.} From each image are extracted several patches of size $k\times k$, which will serve as examples for training the CNN and predicting the latent resolution. In this paper $k = 64$ was used. The extracted patches are centered at the Harris corner locations.

The Algorithm \ref{algo:scnn} is aimed at the multi-class classification version where the latent resolution is discretized into six standard resolutions, and any other resolutions are cast into the nearest of these six resolutions. 
Because the input image or video size is $ m \times 1080$ with $ m \geq 1080$, the largest resolution is $1080$.
The prediction for each patch is the class corresponding to the maximum probability from the $6$-dimensional model output. 
Then, the predicted quality for the whole image is the $90^{th}$ percentile of all the class predictions from the input patches.
\vspace{-4mm}
\begin{algorithm}[h]
\DontPrintSemicolon
\caption{SoftMax-CNN Quality Prediction Algorithm}
\label{algo:scnn}
  \KwInput{Set $K$ of $64 \times 64$ patches from input image $I$}
  \KwOutput{Predicted quality $Q_{SCNN}$}
  \For{patches $\bx_i\in K$}{    
     Compute CNN output $\bp_i=CNN(\bx_i)\in \RR^6$.\\
     Compute prediction $q_i\hspace{-0.5mm}= \hspace{-0.5mm}\argmax(\bp_i)\hspace{-0.5mm}\in\hspace{-0.5mm} \{1,...,6\}$ \\
     }
Obtain the aggregated quality prediction for image \textit{I}
\vspace{-3mm}
\[
   Q_{SCNN} = percentile(\bq,90)
\vspace{-3mm}
\]
\end{algorithm}
\vspace{-9mm}
\subsection{Mask-Based CNN}
\vspace{-1mm}

Unlike conventional neural networks, CNNs effectively employ local receptive fields to extract features from raw data. 
The connection between input and output neurons is performed via convolutions employing trainable kernels, followed by max-pooling layers. 
It was shown \cite{tran2015learning,giannopoulos2018convolutional} that small receptive fields of convolution kernels may lead to higher prediction accuracy than larger kernels.

Fully convolutional networks (FCN) consist only of convolution and max-pooling layers and can obtain an output map for images of arbitrary sizes.
The size of the convolution filters dictates the connection between the input and output size, whether padding was used, and the number and stride of the max pooling layers.
The proposed Mask-CNN from Algorithm \ref{algo:maskcnn} is an FCN where this transformation between the input and the output is explicitly used to propagate a mask of the Harris corners from the input to the output map. 
The mask is propagated by deleting a border for each convolution layer (the size depends on how much padding was used) to obtain an output of the same size as the convolution output and perform max pooling for each max pooling layer.
This way, the correspondents of the Harris corner locations can be identified in the output map, and reliable resolution predictions can be extracted and aggregated into the final resolution prediction.
\begin{algorithm}[h]
\caption{Mask-CNN Quality Prediction Algorithm}
\label{algo:maskcnn}
\DontPrintSemicolon
  \KwInput{Image $I$ of size $m \times n$, corresponding set of corners $K$}
  \KwOutput{Predicted quality, $Q_{MSCNN}$}

  Create a $m \times n$ binary mask $B$ matrix with ones at the corner locations from $K$, otherwise 0. \\
  Compute the $p \times q\times d$ output map $M=CNN(I)$.\\
  Compute the $p \times q$ transformed mask $T$ from $B$ using the induced transformations from \textit{CNN}.\\
  Obtain $S=\{(x,y), T(x,y)\not =0\}$.\\
  \For{$(x_i,y_i)\in S$}{ 
  Compute $q_i\hspace{-0.5mm}= \hspace{-0.5mm}\argmax(M(x_i,y_i))\hspace{-0.5mm}\in\hspace{-0.5mm} \{1,...,d\}$\\
  }
  Obtain the aggregated quality prediction for image \textit{I}
\vspace{-3mm}
\[
   Q_{MSCNN} = percentile(\bq,90)
\vspace{-3mm}
\]
\end{algorithm}
This algorithm has two versions. The classification version, Mask-SoftMax-CNN, has $d=6$ channels in the output map $M$ as shown in the Algorithm \ref{algo:maskcnn}. The regression version (Mask-CNN) has $d=1$. 
For regression, steps 5-6 are removed, and the aggregation step 7 is replaced by $Q_{MCNN} = \frac{1}{|S|}\sum_{(x,y)\in S} M(x,y)$.
\vspace{-4mm}
\subsection{Other ML models}
\vspace{-1mm}

Other ML models were also evaluated, using features obtained by the Mask-CNN Algorithm \ref{algo:maskcnn}. 
For each input image, the features were obtained from the output map $M$ of the Mask-CNN algorithm, with locations from $S$, i.e., $M(S)$.
The values of $M(S)$ were sorted in decreasing order of their value, and the top $50$ values were used as a feature vector to train a multi-class ML model (e.g., Random Forest) for predicting one of the six latent resolutions.

The following ML methods were trained on these features: a decision tree \cite{wu2008top}, a Random Forest \cite{ho1995random} with 300 trees, a Naive Bayes classifier with Gaussian kernel, and a multinomial Logistic Regression classifier.

\vspace{-4mm}
\subsection{CNN Architecture and Training}
\vspace{-1mm}

Both CNNs have four $5 \times 5$ convolution layers, followed by a $8 \times 8$ convolution layer with one filter for the Mask-CNN and six for the SoftMax-CNN and Mask-SoftMax CNN. 
The first two layers have 16 filters, and the following two layers have 32 filters. 
Layers $2$ and $3$ are followed by $2 \times 2$ max-pooling with stride $2$. 
Layer 4 is followed by Rectified Linear Unit (ReLU)\cite{agarap2018deep}. 
Each layer is followed by batch normalization. 

The Mask-CNN was trained with SGD with momentum 0.9 and weight decay $10^{-4}$, while the SoftMax CNN was trained with the Adam optimizer \cite{kingma2014adam}.
The initial batch size was 32 and was doubled every ten epochs. 
The initial learning rate was $10^{-4}$ and was reduced by a factor of 10 when validation performance stopped increasing.

\begin{figure}[ht]
\centering
\includegraphics[width = 8.5cm]{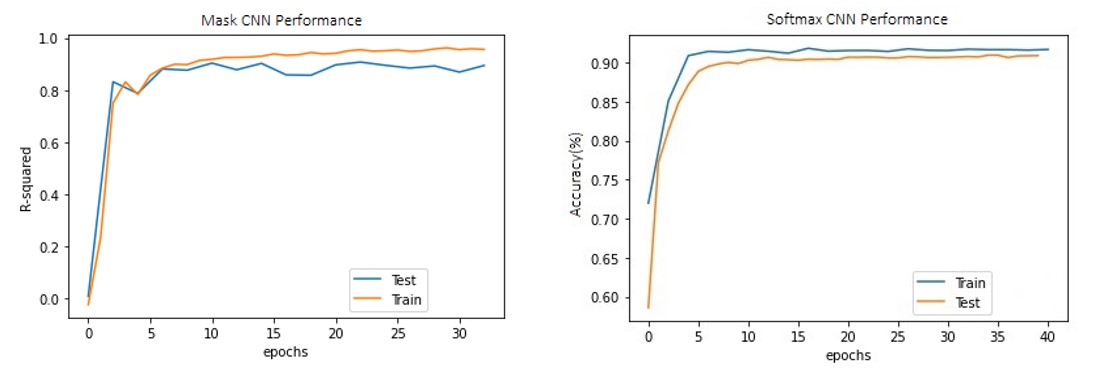}
\vskip -5mm
\caption{Mask CNN $R^2$ (left) and SoftMax CNN accuracy (right) vs epoch number.}
\label{fig:PatchPerform}
\end{figure}

Both models were trained for 40 epochs. 
Figure \ref{fig:PatchPerform} shows the performances on the train and test sets for the two models over the training epochs. 
\vspace{-4mm}

\section{Experiments}
\vspace{-2mm}

Experiments are performed on a dataset specially constructed for the latent resolution problem.

\noindent{\bf Dataset.} 
The full-resolution dataset contains 500 images and 28 videos. The 500 images have at least 1080p resolution and are obtained from \textit{ImageNet} \cite{ILSVRC15}. 
The 28 videos were downloaded from YouTube at 1080p resolution and visually inspected to ensure that they did not have a lower latent resolution. 
Around ten frames were extracted from each video, totaling up to 275 video frames.
The latent resolutions were predicted for the extracted frames, and the 70th percentile was reported as the predicted latent resolution for the whole video.

The 500 images and 28 videos were split into $70\%$ training and $30\%$ testing, obtaining a training set of 350 images and 19 videos and a test set of 150 images and 9 videos.
Training/test images/videos were obtained from the high-resolution training/test images and videos with different latent resolutions as described in Section 2.1 (see Figure \ref{fig:latentres}). 
The exact process was used for classification experiments, where we obtained $9 \times 6 = 54$ test videos. An extra set of 6 videos was downloaded from YouTube with resolutions 240(1), 360(2), 480(1), and 720(2), respectively. 
This way, the classification test set contains 60 videos.

\begin{table}[t]
    \centering
\scalebox{0.9}{
     \begin{tabular}{|l|c|}
   \hline
   Models & $R^2$\\
   \hline
    Mask-CNN &{\bf 0.92}  \\
   \hline
    CNN without mask &0.67 \\
   \hline
    CNN from corner-centered patches &0.91 \\
   \hline
\end{tabular}
}
\vskip -3mm
\caption{Test $R^2$ for different aggregation methods in regression experiments.\\}
\label{tab:regperform}
\vspace{-9mm}
\end{table}


\vspace{-4mm}
\subsection{Regression Experiments}
\vspace{-1mm}

This experiment evaluates the capability of a CNN to predict an arbitrary latent resolution $k\in (0,1)$, as described in Section \ref{sec:latent}. Several aggregation approaches are evaluated.
The first is the Mask-CNN from Algorithm \ref{algo:maskcnn}.
The second one averages all the values from the output mask $M$ to show the importance of mask propagation.
The last one averages the CNN outputs from multiple $64 \times 64$ patches extracted at the corners of the given images to compare it with the mask-based approach that does not need to extract patches.
The results, displayed in Table \ref{tab:regperform} as test $R^2$, show that the mask propagation is essential and the patch-based result is similar to the Mask-CNN result.

\vspace{-4mm}
\subsection{Classification Experiments}
\vspace{-2mm}

The misclassification error in predicting six latent resolutions $\{144,240,360,420,720,1080\}$ was used for the classification task. 
For this task, the classification methods SoftMax CNN and its Mask-SoftMax CNN version, plus the four ML methods (decision tree, RF, Naive Bayes, and logistic regression), were evaluated.  
The test accuracies are shown separately for images and videos in Table \ref{tab:MP}.
\begin{table}[t]
\vspace{-1mm}
    \centering
\scalebox{0.9}{
     \begin{tabular}{|l|c|c|}
   \hline
   &\multicolumn{2}{c|}{Accuracy (\%)}\\
   Models &Images &Videos\\
   \hline
   Naive Bayes & 87.2 & 85\\
   \hline
   Decision Tree & 88.0 & 86.67\\
   \hline
   Random Forest & \textbf{89.6} & 86.67\\
   \hline
   Logistic Regression & 85.1  & \textbf{88.33}\\
   \hline
  Mask-SoftMax CNN & \textbf{97} & \textbf{97}\\
   \hline
   SoftMax CNN & \textbf{95} & \textbf{96} \\
   \hline
\end{tabular}
}
\vskip -3mm
    \caption{Test accuracy for different methods when predicting the latent resolution as one of six classes: 144, 240, 360, 480, 720, and 1080.}
    \label{tab:MP}
\vspace{-6mm}
\end{table}

From Table \ref{tab:MP}, one can see that Mask-SoftMax CNN obtains the highest accuracy, 97\%, followed by its patch-based version SoftMax-CNN with 95\% and Random Forest with 89.6\%. 
The conclusion is that the problem of latent resolution prediction can be solved quite well using Mask-SoftMax CNN, obtaining a test accuracy well above 95\%.

\begin{figure}[ht]
\centering
\includegraphics[width = 8.5cm,height = 2.9cm]{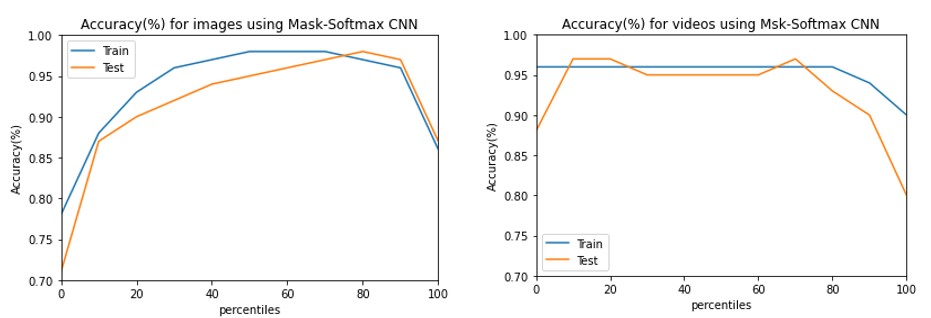}
\vskip -5mm
\caption{Prediction Accuracy(\%) vs. per-frame (left) and per-video (right) aggregation percentiles for Mask-SoftMax CNN.}
\label{fig:percntlVSacc}
\vspace{-5mm}
\end{figure}

Figure \ref{fig:percntlVSacc} shows that Mask-SoftMax CNN is stable to a range of percentiles used to aggregate the per-frame (left) and per-video (right) predictions. The chosen percentiles are 90 per-frame (step 7 of Mask-CNN) and 70 per-video.
\vspace{-4.5mm}

\vspace{-5mm} 
\section{Conclusion}
\vspace{-2mm}

This paper introduced the latent resolution problem, which involves predicting the actual resolution of an image or video to check whether it coincides with the claimed resolution.
It also presented two CNN-based approaches, several Machine Learning approaches to predict the latent resolution and a latent resolution dataset for training and testing/evaluation.
Experiments suggest that the problem can be solved quite well as a regression problem and a multi-class classification for predicting the latent resolution as one or six standard resolutions.
We plan to explore the influence of image downsampling and upsampling methods on the accuracy of latent resolution prediction.
\bibliographystyle{IEEEbib}
\bibliography{references}

\end{document}